%% file: main.tex
\documentclass{article}
\usepackage[T1]{fontenc}
\usepackage{preprint,times}
\input{math_commands.tex}

\usepackage{hyperref}
\usepackage{url}
\usepackage{graphicx}
\usepackage{titlesec}
\usepackage{svg}
\usepackage{float}
\usepackage{wrapfig}
\usepackage{booktabs}
\usepackage{multirow}
\usepackage{makecell}
\usepackage{amsmath,amssymb,amsfonts,bm}
\usepackage{enumitem}
\usepackage{xspace}
\usepackage{microtype}
\usepackage{placeins}
\usepackage{array}
\usepackage{subfigure}
\usepackage{longtable}
\usepackage{algorithm}
\usepackage{algorithmic}
\usepackage{tabularx}
\usepackage{ragged2e}
\usepackage{xcolor} 
\usepackage{pifont} 
\usepackage[most]{tcolorbox}
\hypersetup{hidelinks,bookmarksopen=true}
\newcolumntype{L}[1]{>{\raggedright\arraybackslash}p{#1}}

\newcommand{\benchmarkname}{MechHypoBench}

\newenvironment{takeaway}[1][]
{
\begin{tcolorbox}[
boxrule=0.5pt,
arc=4pt,
left=2pt,
right=2pt,
bottom=2pt,
top=2pt,
rounded corners]
\textbf{#1.}
\small\itshape
}
{
\end{tcolorbox}
}

\title{From Scientific Observations to Mechanisms: Benchmarking Hypothesis Generation by AI Scientists}

\input{arxiv_authors.tex}
\preprintfinalcopy

\begin{document}
\maketitle
\fancyhead{}
\lhead{Preprint. Under review.}

\begin{abstract}
Data-driven mechanistic hypotheses are essential to scientific discovery because they explain how underlying processes produce observed phenomena. AI agents and AI scientists increasingly support scientific data analysis. However, their ability to turn empirical findings into mechanistic hypotheses remains insufficiently examined. To address this gap, we introduce MechHypoBench, the first benchmark for evaluating whether AI agents and AI scientists can generate such hypotheses from empirical data. It combines paper-derived mechanisms from 14 scientific fields with real-world datasets containing 17.98 million records. The construction retains the observational complexity of empirical data while providing a specified underlying mechanism. Agents analyze the observations and propose open-form hypotheses. We develop an evaluation framework that assesses open-form mechanistic hypotheses through their consequences under withheld conditions. Experiments with general agents and AI scientists reveal a substantial gap between generated hypotheses and the underlying mechanisms. 

\end{abstract}

\input{chapters/0_intro}
\input{chapters/1_related}
\input{chapters/2_benchmark}
\input{chapters/3_experiments}
\input{chapters/4_conclusion}

\bibliography{cite,systems}
\bibliographystyle{plainnat}
\end{document}

%% file: math_commands.tex
\usepackage{amsmath,amsfonts,bm}

\def\eqref#1{equation~\ref{#1}}

\def\1{\bm{1}}

\DeclareMathAlphabet{\mathsfit}{\encodingdefault}{\sfdefault}{m}{sl}
\SetMathAlphabet{\mathsfit}{bold}{\encodingdefault}{\sfdefault}{bx}{n}



%% file: arxiv_authors.tex
\author{
Xiaxun Xie$^{1}$ \quad Qingqing Long$^{1,\dagger}$ \quad Meng Xiao$^{2}$ \quad Wei Ju$^{3}$ \\Yuanchun Zhou$^{1}$ \quad Hengshu Zhu$^{1}$ \quad Xuezhi Wang$^{1,\dagger}$ \\[6pt]
$^{1}$Computer Network Information Center, Chinese Academy of Sciences\\
$^{2}$National University of Singapore \\
$^{3}$Sichuan University \\[3pt]
$^{\dagger}$Corresponding authors
}

%% file: chapters/0_intro.tex
\section{Introduction}
\label{sec:introduction}

\begingroup
\hyphenpenalty=1500
\emergencystretch=1em
\setlength{\columnsep}{16pt}
\begin{wrapfigure}{r}{0.34\linewidth}
  \centering
  \setlength{\abovecaptionskip}{3pt}
  \includegraphics[width=\linewidth]{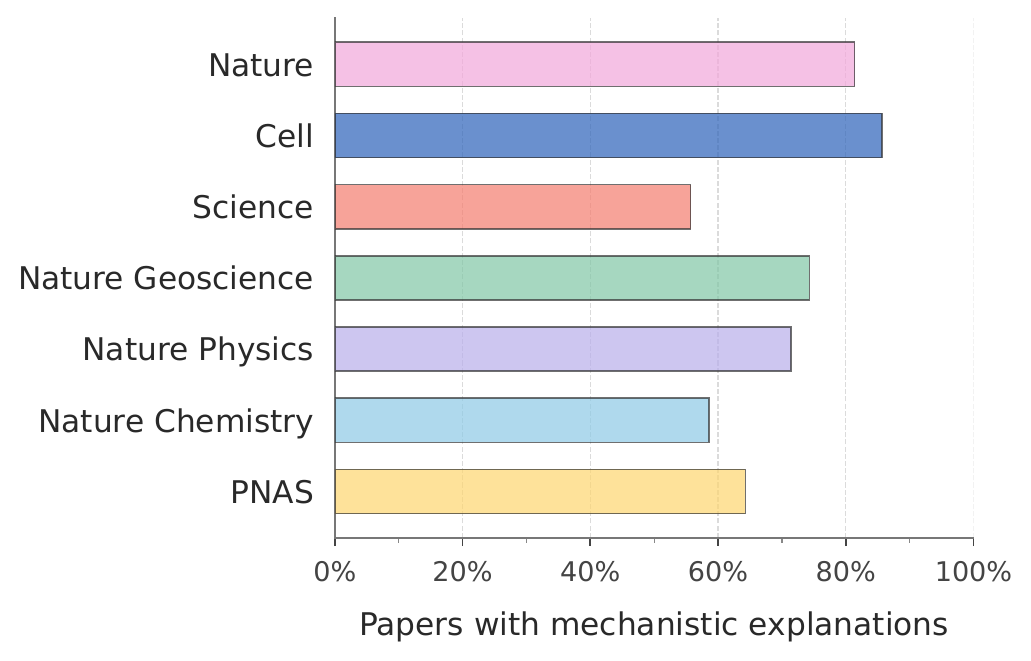}
  \vspace{-6mm}
  \caption{We randomly select papers published in 2021-2025 from high-impact journals. Overall, 70.2\% of the papers treated mechanisms as explanatory outputs, motivating our research on mechanism-driven hypothesis generation.}
  \label{fig:motivation-audit}
\end{wrapfigure}

Recent advances in AI agents and AI scientists have largely accelerated scientific discoveries across domains~\citep{wang2023scientific,boiko2023autonomous,yamada2025ai,gottweis2026accelerating}, including physics, biology, chemistry, etc. One of the central tasks is to use observed data to investigate the underlying mechanism of how scientific phenomena arise. For example, when a catalytic reaction slows over time, researchers may hypothesize that molecules progressively block the catalyst's active sites, leaving fewer sites available for reaction. A mechanistic hypothesis is an open-form account of how processes and their interactions produce an observed phenomenon, including the conditions under which they operate. In our review of papers from natural-science journals, 70.2\% treated a mechanism as an explanatory output (Figure~\ref{fig:motivation-audit}). Assessing AI agents’ ability to generate mechanistic hypotheses from data is therefore important for determining whether they can contribute to scientific discovery.
\par\endgroup

\begin{figure}[!htbp]
  \centering
  \includegraphics[width=0.96\linewidth]{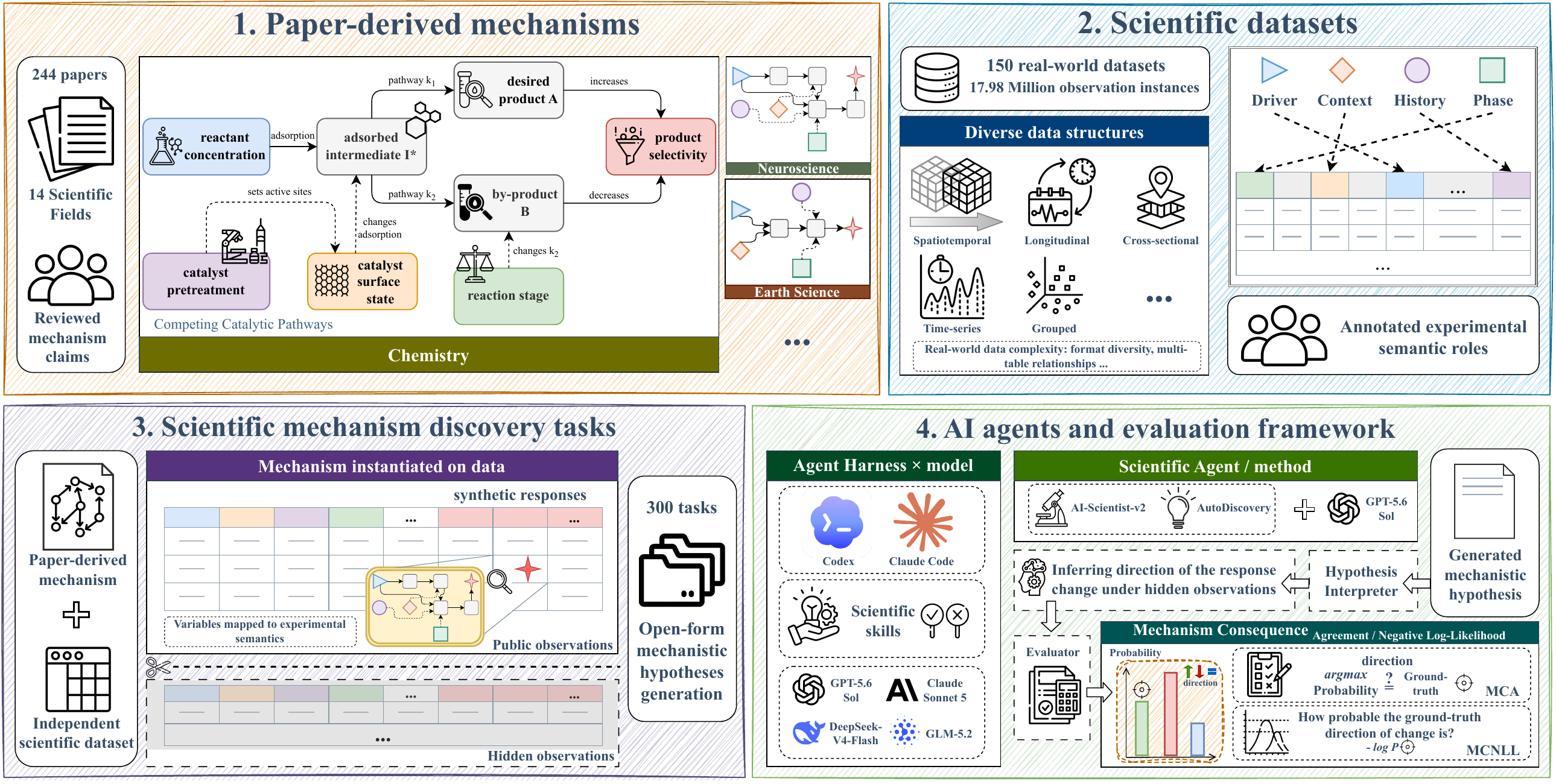}
  \vspace{-2mm}
\caption{Overview of our constructed benchmark.}
 \vspace{-6mm}
  \label{fig:benchmark-overview}
\end{figure}

Existing benchmarks have expanded the evaluation of LLMs in scientific discovery beyond question answering~\citep{jin2019pubmedqa,wang2023scibench}. Several benchmarks study literature-grounded hypothesis and idea generation using scientific papers and background knowledge~\citep{guo2025ideabench,liu2026researchbench,lew2026projectionbench}. Others evaluate hypothesis discovery from specified entities or scientific data, including the recovery of entity relations, equations, data relationships, explanatory patterns, and admissible hypothesis sets~\citep{xiong2025toward,majumder2025discoverybench,shojaee2025llm,liu2025hypobench,chen2026hypospace}. A separate group evaluates scientific research and discovery workflows, ranging from executable data-analysis to broader task- and project-level evaluations~\citep{chen2025scienceagentbench,bragg2026astabench,song2025evaluating}.

However, existing benchmarks pose unique challenges for evaluating data-driven mechanistic hypotheses. 
\textbf{(1) Finding a pattern does not explain how a phenomenon arises.}
Most data-driven benchmarks ask models to recover relationships or equations~\citep{majumder2025discoverybench,shojaee2025llm}. Such tasks show whether a model can detect what happens in the data, but not whether it can explain how underlying processes produce the phenomenon under specific conditions. The \textit{Mechanistic} column of Table~\ref{tab:related-benchmarks} shows that direct evaluation of this ability remains limited, even among benchmarks that accept explanatory hypotheses.
\textbf{(2) Known mechanisms rarely come with realistic observations.}
Real datasets preserve the complexity of scientific observations, but they rarely identify a unique and fully specified mechanism~\citep{liu2025hypobench}. Treating a source-paper explanation as ground truth may privilege one interpretation and reward familiarity with the paper rather than reasoning from the data. Synthetic data provide exact references but simplify the observations~\citep{chen2026hypospace}. Accordingly, the \textit{Exp. Data} and \textit{Source} columns of Table~\ref{tab:related-benchmarks} show that no prior benchmark fully combines realistic empirical data with a controllable underlying mechanism.
\textbf{(3) Open-form hypotheses cannot be scored by wording alone.}
Equivalent mechanisms may use different wording, while a hypothesis may have clear implications only under some conditions. Text matching or a single classification decision cannot capture these differences. Evaluation should instead test the outcomes implied under unseen conditions to assess how well a hypothesis captures the behavior of the underlying mechanism. The pipeline should also support agents that iteratively inspect data and execute analyses. The \textit{Open-form} and \textit{Agent} columns of Table~\ref{tab:related-benchmarks} show that existing benchmarks rarely provide both.

In response, we introduce \benchmarkname, a benchmark for data-driven mechanistic hypothesis generation with AI agents (Figure~\ref{fig:benchmark-overview}). We instantiate mechanisms derived from papers spanning 14 scientific fields~\citep{priem2022openalex}  on independently collected real-world scientific datasets. The datasets vary in structure and collectively contain 17.98 million observation records. This construction combines a scientifically grounded generating process with empirical observational complexity by aligning mechanisms and datasets through their experimental roles. Agents analyze the resulting observations, then propose open-form mechanistic hypotheses to explain how the observed phenomena arise. We assess these hypotheses through their implied consequences under withheld changes in conditions, using ground truth outcomes determined by the same generating mechanism. Our experiments reveal substantial room for improvement in agents’ ability to formulate mechanistic hypotheses from scientific data. \textbf{We summarize our contributions as follows:}
\begin{itemize}[nosep,leftmargin=*]
\setlength{\topsep}{-3pt}
\setlength{\itemsep}{0pt}
\setlength{\parsep}{0pt}
    \item We propose \benchmarkname, the first benchmark for data-driven mechanistic hypothesis generation. It combines paper-derived mechanisms from 14 scientific fields with independently collected empirical datasets.
    \item We develop an evaluation framework for open-form mechanistic hypotheses. It assesses how well a hypothesis captures the behavior of the underlying mechanism.
    \item Through experiments with AI scientists, we identify limitations in data-driven mechanistic hypothesis generation. A substantial gap remains between the generated hypotheses and the underlying mechanisms.
\end{itemize}

%% file: chapters/1_related.tex
\section{Related Work}
\label{sec:related}
We position \benchmarkname\ among existing benchmarks for scientific reasoning and discovery in Table~\ref{tab:related-benchmarks}. These benchmarks extend beyond question answering~\citep{jin2019pubmedqa,wang2023scibench}. Literature- and knowledge-driven benchmarks evaluate idea or hypothesis generation from published research and background knowledge~\citep{guo2025ideabench,liu2026researchbench,lew2026projectionbench,xiong2025toward}. But discovery also requires direct analysis of observational data to establish empirical support for these proposals.
Data-driven benchmarks can be broadly divided into two directions. One line focuses on recovering structured scientific targets, ranging from observable relationships and equations~\citep{shojaee2025llm}, to enumerable hypothesis spaces~\citep{chen2026hypospace}, and structured mechanistic models in MechBench~\citep{yu2026mechbench}. Another line explores open-form hypothesis generation from scientific datasets or analysis contexts~\citep{majumder2025discoverybench,liu2025hypobench}.

However, existing empirical settings generally lack explicit grounding in fully specified underlying mechanisms. Controlled synthetic settings, in turn, simplify the complexity of real scientific observations. 
The growing use of AI agents also motivates evaluating how they conduct analyses and interact with tools in discovery. Research-workflow-based benchmarks focus on evaluating executable analyses, tool usage, and end-to-end scientific tasks~\citep{chen2025scienceagentbench,bragg2026astabench,song2025evaluating}. 
More recent efforts, such as EurekaBench~\citep{geng2026eurekabenchmeasuringagenticability}, further explore autonomous scientific research through iterative experimentation and mechanism discovery. 
These works either emphasize workflow execution or rely on interactive environments with predefined discovery processes, leaving open the question of whether AI agents can directly formulate open-form mechanistic explanations from complex empirical observations. \benchmarkname\ addresses this gap by evaluating open-form mechanistic hypothesis generation from empirical scientific datasets while preserving their observational complexity.

\input{chapters/related_benchmarks.tex}

%% file: chapters/related_benchmarks.tex
\setlength{\textfloatsep}{5pt}
\begin{table}[htbp]
\caption{Comparison of \benchmarkname\ with existing scientific discovery benchmarks. $\sim$ denotes partial coverage. \# Sci. Subj. denotes the number of real scientific fields covered by each work. Emp. Data indicates whether tasks are constructed from empirical scientific datasets.}
\label{tab:related-benchmarks}
\centering
\begingroup
\scriptsize
\setlength{\tabcolsep}{1.5pt}
\renewcommand{\arraystretch}{1.04}
\setlength{\extrarowheight}{1pt}
\setlength{\heavyrulewidth}{0.6pt}
\setlength{\lightrulewidth}{0.25pt}
\setlength{\aboverulesep}{1.5pt}
\setlength{\belowrulesep}{1.5pt}
\setlength{\arrayrulewidth}{0.25pt}

\newlength{\benchcontentwidth}
\setlength{\benchcontentwidth}{\dimexpr\linewidth-18\tabcolsep\relax}

\newcommand{\benchpartial}{\ensuremath{\sim}}

\begin{tabular}{@{}
>{\centering\arraybackslash}m{0.085\benchcontentwidth}
>{\centering\arraybackslash}m{0.300\benchcontentwidth}
>{\centering\arraybackslash}m{0.070\benchcontentwidth}
>{\centering\arraybackslash}m{0.065\benchcontentwidth}
>{\centering\arraybackslash}m{0.115\benchcontentwidth}
>{\centering\arraybackslash}m{0.100\benchcontentwidth}
>{\centering\arraybackslash}m{0.090\benchcontentwidth}
>{\centering\arraybackslash}m{0.045\benchcontentwidth}
>{\centering\arraybackslash}m{0.050\benchcontentwidth}
>{\centering\arraybackslash}m{0.085\benchcontentwidth}@{}}
\toprule
\multirow{2}{*}{\textbf{Type}}
& \multirow{2}{*}{\textbf{Work}}
& \multirow{2}{*}{\shortstack{\textbf{\# Sci.}\\\textbf{Subj.}}}
& \multicolumn{2}{c}{\textbf{Task Source}}
& \multicolumn{2}{c}{\textbf{Hypothesis}}
& \multicolumn{2}{c}{\textbf{Agent Pipeline}}
& \multirow{2}{*}{\shortstack{\textbf{Task}\\\textbf{Scale}}} \\
\cmidrule(lr){4-5}
\cmidrule(lr){6-7}
\cmidrule(lr){8-9}
&
&
& \shortstack{\textbf{Emp. Data}}
& \shortstack{\textbf{Source}}
& \shortstack{\textbf{Mechanistic}}
& \shortstack{\textbf{Open-form}}
& \textbf{Skill}
& \textbf{Agent}
&
\\
\midrule

\multirow{4}{*}{\shortstack{Literature\\-Driven}}
& IdeaBench~\citep{guo2025ideabench}
& 1
& \textcolor{red}{\ding{55}} & papers
& \textcolor{red}{\ding{55}} & \textcolor{green}{\ding{51}}
& \textcolor{red}{\ding{55}} & \textcolor{red}{\ding{55}}
& unknown \\
\noalign{\vskip 2pt}
& ResearchBench~\citep{liu2026researchbench}
& 12
& \textcolor{red}{\ding{55}} & papers
& \textcolor{red}{\ding{55}} & \textcolor{green}{\ding{51}}
& \textcolor{red}{\ding{55}} & \textcolor{red}{\ding{55}}
& unknown \\
\noalign{\vskip 2pt}
& ProjectionBench~\citep{lew2026projectionbench}
& 1
& \textcolor{red}{\ding{55}} & papers
& \textcolor{red}{\ding{55}} & \textcolor{green}{\ding{51}}
& \textcolor{red}{\ding{55}} & \textcolor{red}{\ding{55}}
& unknown \\
\noalign{\vskip 2pt}
& TruthHypo~\citep{xiong2025toward}
& 1
& \textcolor{red}{\ding{55}} & Bio. KG
& \textcolor{red}{\ding{55}} & \textcolor{red}{\ding{55}}
& \textcolor{red}{\ding{55}} & \textcolor{red}{\ding{55}}
& 2k tasks \\
\midrule

\multirow{6}{*}{\shortstack{Research\\Workflow}}
& ScienceAgentBench~\citep{chen2025scienceagentbench}
& 4
& \textcolor{green}{\ding{51}} & 44 (paper, dataset, code)
& \textcolor{red}{\ding{55}} & \textcolor{red}{\ding{55}}
& \textcolor{red}{\ding{55}} & \textcolor{green}{\ding{51}}
& 102 tasks \\
\noalign{\vskip 2pt}
& AstaBench~\citep{bragg2026astabench}
& unknown
& \benchpartial & 11 benchmarks
& \textcolor{red}{\ding{55}} & \benchpartial
& \textcolor{red}{\ding{55}} & \textcolor{green}{\ding{51}}
& 2.4k tasks \\
\noalign{\vskip 2pt}
& SDE~\citep{song2025evaluating}
& 4
& \benchpartial & 43 scenarios
& \benchpartial & \textcolor{red}{\ding{55}}
& \textcolor{red}{\ding{55}} & \textcolor{red}{\ding{55}}
& 1k QA + 8 proj. \\
\noalign{\vskip 2pt}
& EurekaBench~\citep{geng2026eurekabenchmeasuringagenticability}
& 6
& \benchpartial & simulated environments
& \textcolor{green}{\ding{51}} & \benchpartial
& \textcolor{red}{\ding{55}} & \textcolor{green}{\ding{51}}
& 26 tasks \\
\midrule

\multirow{4}{*}{\shortstack{Data-Driven\\Structured\\Recovery}}
& LLM-SRBench~\citep{shojaee2025llm}
& 4
& \textcolor{red}{\ding{55}} & equations
& \textcolor{green}{\ding{51}} & \textcolor{red}{\ding{55}}
& \textcolor{red}{\ding{55}} & \textcolor{red}{\ding{55}}
& 239 tasks \\

\noalign{\vskip 2pt}
& HypoSpace~\citep{chen2026hypospace}
& unknown
& \textcolor{red}{\ding{55}} & synth.
& \textcolor{red}{\ding{55}} & \textcolor{red}{\ding{55}}
& \textcolor{red}{\ding{55}} & \textcolor{red}{\ding{55}}
& unknown \\

\noalign{\vskip 2pt}
& MechBench~\citep{yu2026mechbench}
& 4
& \textcolor{red}{\ding{55}} & canonical mechanisms
& \textcolor{green}{\ding{51}} & \textcolor{red}{\ding{55}}
& \textcolor{red}{\ding{55}} & \textcolor{green}{\ding{51}}
& 512 tasks \\
\midrule

\multirow{4}{*}{\shortstack{Data-Driven\\Hyp. Gen.}}
& DiscoveryBench~\citep{majumder2025discoverybench}
& 6
& \benchpartial & 27 (paper, dataset, code)
& \textcolor{red}{\ding{55}} & \textcolor{green}{\ding{51}}
& \textcolor{red}{\ding{55}} & \textcolor{green}{\ding{51}}
& 264 real +\ 903 synth.\\

\noalign{\vskip 2pt}
& HypoBench~\citep{liu2025hypobench}
& 7
& \benchpartial & 194 datasets
& \benchpartial & \textcolor{green}{\ding{51}}
& \textcolor{red}{\ding{55}} & \textcolor{red}{\ding{55}}
& unknown \\

\cline{2-10}
& \textbf{Ours}
& \textbf{14}
& \textcolor{green}{\ding{51}}
& \textbf{244 papers +}\newline\textbf{150 datasets}
& \textcolor{green}{\ding{51}}
& \textcolor{green}{\ding{51}}
& \textcolor{green}{\ding{51}}
& \textcolor{green}{\ding{51}}
& \textbf{300 tasks} \\
\bottomrule
\end{tabular}
\par\smallskip
\endgroup
\end{table}

%% file: chapters/2_benchmark.tex
\titlespacing{\subsection}
{0pt}{4pt}{2pt}

\section{MechHypoBench}
\label{sec:benchmark}

\subsection{Concept Formalization}
\label{sec:formulization}

\noindent\textbf{Mechanisms, conditions, and observations.}\quad Scientific inquiry draws on observed data collected under particular conditions to investigate how phenomena arise~\citep{woodward2011data}. An underlying mechanism \(\mathcal M\) generates the phenomenon, but the available observations may reveal only part of how it operates. Its observable response may depend not only on the factor being examined, but also on the surrounding context, prior exposure, and the stage of the process~\citep{machamer2000thinking}. To represent these dependencies, we distinguish four \textbf{experimental semantic roles}: 
\begin{itemize}[nosep,leftmargin=*]
	\item \textbf{Driver (\(\mathrm d\)):} the factor whose effect is examined.
	\item \textbf{Context (\(\mathrm x\)):} a contextual factor that may modify the driver's effect.
	\item \textbf{History (\(\mathrm h\)):} prior exposure or state.
	\item \textbf{Phase (\(\mathrm p\)):} the stage of the process.
\end{itemize}
A condition \(c=(c_{\mathrm d},c_{\mathrm x},c_{\mathrm h},c_{\mathrm p})\in\mathcal C\) specifies the joint values or states associated with these roles, where \(\mathcal C\) denotes the relevant condition space. Let \(R_{\mathcal M}(c)\) denote the response, the observable outcome generated by mechanism \(\mathcal M\) under condition \(c\). For instance, in the chemistry example of Figure~\ref{fig:benchmark-overview}, \(c_{\mathrm d}\) specifies the \emph{reactant concentration}, \(c_{\mathrm x}\) the \emph{catalyst surface state}, \(c_{\mathrm h}\) the \emph{catalyst pretreatment}, and \(c_{\mathrm p}\) the \emph{reaction stage}. Together, these specify a condition \(c\), with \emph{product selectivity} as the response \(R_{\mathcal M}(c)\). Changing the \emph{reactant concentration} \(c_{\mathrm d}\) while keeping the other three components fixed yields a different condition. Let \(\mathcal C^\mathrm{obs}\subseteq\mathcal C\) denote the conditions under which observations are available. These observations can be represented as condition-response pairs: \(\mathcal O_{\mathcal M}(\mathcal C^\mathrm{obs})=\{(c,R_{\mathcal M}(c))\mid c\in\mathcal C^\mathrm{obs}\}.\) In the chemistry example, \(\mathcal C^\mathrm{obs}\) contains the combinations of \emph{concentration, catalyst state, pretreatment}, and \emph{reaction stage} under which measurements were collected; \(\mathcal O_{\mathcal M}(\mathcal C^\mathrm{obs})\) pairs those combinations with their corresponding \emph{product selectivity}.

\noindent\textbf{Mechanistic hypothesis.}\quad Since the available observations do not fully reveal how the underlying mechanism operates, researchers formulate mechanistic hypotheses to explain how the observed phenomenon arises~\citep{darden2002strategies,craver2013search}. A mechanistic hypothesis \(h\) is an open-form account of how relevant processes and their interactions produce the observations \(\mathcal O_{\mathcal M}(\mathcal C^{\mathrm{obs}})\), including the conditions under which they operate. The available observations may support multiple mechanistic hypotheses to differing degrees.

\subsection{Scientific Source Collection}
\label{sec:collection}

We collect scientific papers to derive mechanisms and independently gather real-world empirical datasets on which these mechanisms are instantiated.

\noindent\textbf{Paper collection.}\quad We collected candidate papers through Europe PMC and OpenAlex\citep{priem2022openalex} across a range of scientific journals, including PLOS journals and eLife. \textbf{(1) Candidate filtering.} We retained empirical studies that proposed, tested, or revised mechanisms, excluding reviews, protocols, retracted papers, and studies focused solely on prediction or optimization. \textbf{(2) Mechanistic claim review.} Human review assessed whether the reported claims described how a phenomenon arises and provided sufficient information to identify the relevant experimental semantic roles and conditions. \textbf{(3) Claim extraction and verification.} We used LLMs to assist in extracting mechanistic claims and supporting passages, then manually checked the extracted content against the source papers, preserving the claims and their conditions of applicability. The resulting collection contains 306 papers spanning 14 scientific fields.

\noindent\textbf{Dataset collection.}\quad We independently collected candidate datasets from scientific repositories, including PANGAEA~\citep{felden2023pangaea},  ScienceDB~\citep{long2026sciencedb}, Zenodo~\citep{zenodo}, Dryad~\citep{dryad}, and the Open Science Framework~\citep{foster2017open}. \textbf{(1) Candidate filtering.} We selected datasets with traceable provenance, accessible data files, suitable reuse licenses, and documentation describing the recorded variables. \textbf{(2) Scientific and structural review.} Human review assessed whether each dataset contained sufficient variation, ordered observations, auxiliary measurements, and grouping structure to support the experimental semantic roles and the intended task construction. \textbf{(3) Dataset preparation and role annotation.} Dataset variables were manually annotated with their corresponding experimental semantic roles. Dataset-specific preprocessing adapters accommodated differences in file formats and table layouts. The resulting collection contains 200 datasets with manual role annotations and diverse structures, including cross-sectional, longitudinal, time-series, grouped, and spatiotemporal formats, among others.

\subsection{Scientific Skills Collection}
\label{sec:skills}

To examine whether scientific skills affects mechanistic hypothesis generation, we collected over 500,000 science-related skills from GitHub and filtered them for relevance to scientific data analysis, discovery, and hypothesis generation. \textbf{(1) Initial filtering.} An LLM used for screening skill descriptions to shortlist over 20,000 candidates. Filtering excluded academic writing and formatting tools, literature-retrieval and web-dependent tools, and generic agent frameworks or LLM API wrappers. \textbf{(2) Relevance review.} Further LLM-based review examined both descriptions and skill contents to identify candidates relevant to scientific data analysis, discovery, or hypothesis generation. \textbf{(3) Pool consolidation.} After removing duplicates and entries with insufficient content, the final pool contained 4,043 skills.

\subsection{Task Construction}
\label{sec:construction}

\begin{figure}[ht]
\centering
\includegraphics[width=\linewidth]{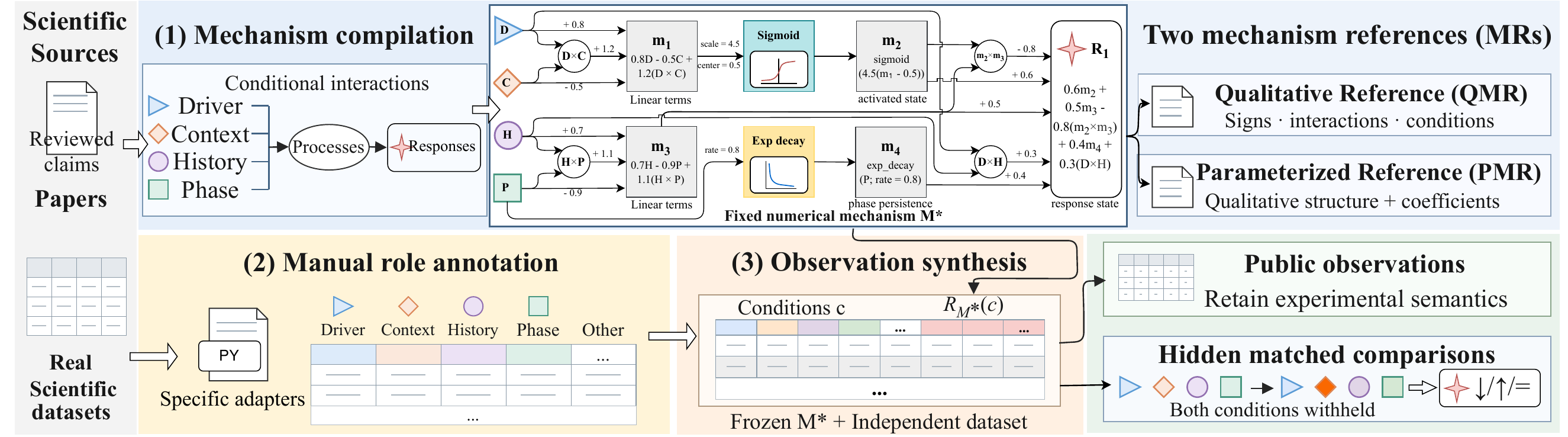}
\vspace{-3mm}
\caption{Task construction pipeline. Paper-derived claims are compiled into a fixed mechanism \(\mathcal M^*\) and aligned with independent scientific datasets through experimental roles. The mechanism generates responses on the aligned conditions and yields two open-form references. Agents receive public observation and are evaluated on hidden matched comparisons.}
\label{fig:task-construction}
\end{figure}

Figure~\ref{fig:task-construction} illustrates how we simulate the scientific inference setting using a specified \textbf{paper-derived mechanism} \(\mathcal M^*\), which serves as the underlying mechanism within each task. It is known to the benchmark designer but hidden from the agent. We use LLM-assisted conversion to organize reviewed paper claims into a directed graph of processes, interactions, and responses associated with the experimental semantic roles. Numerical functions and parameters specify how these components operate. The resulting mechanism is checked against the reviewed claims and fixed before dataset pairing, providing a controlled realization of the reported processes rather than a numerical reconstruction of the original experiment. From the same mechanism graph, we derive two open-form textual \textbf{mechanism references} (MRs): the \textbf{qualitative mechanism reference} (QMR) describes its dependencies, interactions, and conditions, while the \textbf{parameterized mechanism reference} (PMR) additionally provides the corresponding equations and numerical parameters.

We align an independently collected dataset with \(\mathcal M^*\) through its manually annotated experimental semantic roles. Dataset variables specify conditions \(c\), under which the mechanism generates responses \(R_{\mathcal M^*}(c)\). This preserves empirical sampling patterns, grouping, and auxiliary measurements, with added noise representing measurement variability. Dataset labels and task descriptions retain the experimental roles and their interpretation while replacing source-specific terminology to avoid misleading domain associations and cues to memorized scientific accounts. The generated observations are divided into \textbf{public observations} \(\mathcal O_{\mathcal M^*}(\mathcal C^{\mathrm{obs}})\), available to the agent, and \textbf{hidden observations}, withheld for assessment. Each task includes a set \(\mathcal Q\) of \textbf{matched comparisons} \(q=(c^{(1)},c^{(0)},R)\), each comparing response \(R\) under two conditions that differ in exactly one component. These comparisons have stable ground-truth direction of response changes determined by \(\mathcal M^*\) and sufficient observational support. Observations under both compared conditions belong to the hidden portion, although individual component values may still occur in the public observations. Task difficulty is determined by a marginal baseline fitted to the public observations: easy tasks yield one to three errors across ten directional comparisons, whereas hard tasks yield six to ten, distinguishing how poorly marginal relationships capture the conditional response changes.

\subsection{Human Evaluation}
\label{sec:human-evaluation}

To assess the quality of the constructed tasks, we conducted a human evaluation across both difficulty levels. Ten reviewers with scientific backgrounds assessed each task on four dimensions using a 0-5 scale: (1)  \textbf{Scientific Coherence} (SC) assesses whether the mechanism, experimental roles, and comparisons form a scientifically interpretable scenario; (2) \textbf{Mechanism-task Consistency} (MC) measures agreement among the mechanism, role mapping, data-generation process, and comparisons; (3) \textbf{Probe Clarity} (PC) assesses whether the condition change, response variable, and requested judgment are unambiguous; and (4) \textbf{Reference-Answer Validity} (RV) assesses how clearly the ground-truth direction of response changes are supported by the specified mechanism.

\begin{wraptable}[10]{r}{0.38\columnwidth}
  \centering
  \vspace{-5pt}
  \caption{Human evaluation scores after mechanism reconstruction (0-5).}
  \label{tab:human-evaluation}
  \small
  \setlength{\tabcolsep}{4pt}
  \renewcommand{\arraystretch}{1.4}
  \begin{tabular*}{\linewidth}{@{\extracolsep{\fill}}lcccc@{}}
    \toprule
    \textbf{Task} & \textbf{SC} & \textbf{MC}
    & \textbf{PC} & \textbf{RV} \\
    \midrule
    Easy & 4.62 & 4.69 & 4.90 & 4.97 \\
    Hard & 4.60 & 4.54 & 4.90 & 4.96 \\
    \bottomrule
  \end{tabular*}
\end{wraptable}

Human review prompted the reconstruction of 17 paper-derived mechanisms with insufficient scientific coherence.  Table~\ref{tab:human-evaluation} reports scores after these revisions. High ratings for probe clarity (4.90) and reference-answer validity (4.96-4.97) support the clarity of the comparisons and the validity of their ground-truth response changes. Scientific coherence (4.60-4.62) and mechanism-task consistency (4.54-4.69) indicate that the revised tasks present interpretable scientific scenarios with internally consistent construction.

\subsection{Evaluation Framework}
\label{sec:evaluation}

\noindent\textbf{Evaluation Framework.}\quad Mechanistic hypotheses support reasoning about interventions and outcomes under conditions not yet observed~\citep{woodward2005making,craver2007explaining}. Motivated by this perspective, we assess hypotheses through the response changes they imply for the hidden matched comparisons. For each \(q=(c^{(1)},c^{(0)},R)\), let \(\Delta_{\mathcal M^*}(q)\) denote the ground-truth direction of change in response \(R\) from \(c^{(0)}\) to \(c^{(1)}\) determined by the paper-derived mechanism, and let \(\Delta_h(q)\) denote the corresponding direction implied by hypothesis \(h\). Agreement between these directions provides a criterion for assessing the hypothesis. Open-form mechanistic hypotheses do not necessarily specify an executable response function, so their implications must be interpreted before scoring. Asking the generating agent to answer the comparisons could draw on observations or intermediate analyses not expressed in its hypothesis. We therefore use a separate, fixed LLM-based \textbf{hypotheses interpreter} that receives only \(h\) and the comparison descriptions, without access to the public observations or ground-truth directions. For each comparison \(q\), it returns \(p_h(t\mid q)\) over \(\mathcal T=\{\mathrm{decrease},\mathrm{no\_change},\mathrm{increase}\}\), representing its interpretation of \(\Delta_h(q)\). We request probabilities rather than a single direction because a hypothesis may leave some consequences ambiguous or incompletely specified. A probability distribution allows the interpreter to express this uncertainty instead of forcing an equally definite answer for every hypothesis. A deterministic evaluator then computes scores against \(\Delta_{\mathcal M^*}(q)\). The same interpreter processes both MRs, providing a comparison with specified underlying mechanism content in each task. Empty and shuffled controls provide baselines for absent and mismatched content, respectively.

\noindent\textbf{Metrics.}\quad
We assess each mechanistic hypothesis using two complementary metrics:
\nopagebreak[4]
\[
\renewcommand{\arraystretch}{1.4}
\begin{array}{@{}l@{\;}l@{}}
\operatorname{MCA}(h)
&\displaystyle =\frac{1}{|\mathcal Q|}\sum_{q\in\mathcal Q}
\mathbf{1}\!\left[
\arg\max_{t\in\mathcal T}p_h(t\mid q)
=\Delta_{\mathcal M^*}(q)
\right],\\
\operatorname{MCNLL}(h)
&\displaystyle =-\frac{1}{|\mathcal Q|}\sum_{q\in\mathcal Q}
\log p_h\!\left(\Delta_{\mathcal M^*}(q)\mid q\right).
\end{array}
\]
Mechanism Consequence Agreement (MCA) evaluates how consistently the hypothesis-implied direction of response change agree with those determined by the generating mechanism. Mechanism Consequence Negative Log-Likelihood (MCNLL) further evaluates the probability assigned to the ground-truth direction, penalizing confident errors more strongly. Together, higher MCA and lower MCNLL provide evidence that a mechanistic hypothesis captures the behavior of the underlying mechanism under the tested conditions.

%% file: chapters/3_experiments.tex
\titlespacing{\subsection}
{0pt}{4pt}{2pt}

\section{Experiments}
\label{sec:experiments}

\subsection{Experimental Setup}
\label{sec:setup}

\noindent\textbf{Tasks.}\quad Our evaluation includes 300 tasks, drawing on 244 papers and 150 datasets from source collections of 306 papers and 200 datasets (Section~\ref{sec:collection}). It comprises 120 easy and 180 hard tasks.

\noindent\textbf{Evaluated AI scientists and Agents.} (1) \textbf{General Agents.} General AI agents are increasingly being used to assist and accelerate scientific discovery. We evaluate Codex~\citep{openai2026codexcli} with GPT-5.6 Sol~\citep{openai2026sol} and Claude Code~\citep{anthropic2026claudecode} with Claude Sonnet 5~\citep{anthropic2026sonnet5}. To compare the two harnesses with shared model backends, we also pair each with DeepSeek-V4-Flash~\citep{deepseekai2026deepseekv4} and GLM-5.2~\citep{glm5team2026glm5vibecodingagentic}. Each of the six harness-model configurations is evaluated with and without scientific skills.
(2) \textbf{AI Scientists.} To test whether workflows designed for scientific discovery help with mechanistic hypothesis generation from observational data, we evaluate the ideation component of AI-Scientist-v2~\citep{yamada2025ai} and the Bayesian-surprise-driven search procedure of AutoDiscovery~\citep{agarwal2026autodiscovery}.

\noindent\textbf{Implementation details.}\quad For skill-enabled runs, SkillRouter-Reranker-0.6B~\citep{zheng2026skillrouterskillroutingllm} selects 32 skills from the collected pool (Section~\ref{sec:skills}) using the task description and public dataset information, including its overview, structure, and format. General-agent runs use a 30-minute terminal budget and high reasoning settings. Across the twelve general-agent configurations, 214-289 of the 300 tasks yield valid terminal outputs. Invalid or missing outputs are scored using the result from the empty-control condition. Both AI-scientist systems use GPT-5.6 Sol. AutoDiscovery is evaluated on a fixed, stratified subset of 40 tasks, comprising 20 easy and 20 hard, since its search used 4,464 accepted LLM calls, averaging 111.6 per task. GPT-5.6 Terra~\citep{openai2026terra}, with high reasoning and temperature zero, serves as the fixed hypothesis interpreter. Details of interpretation and scoring are given in Section~\ref{sec:evaluation}.

\input{chapters/main_results}

\subsection{Evaluating the Ability of Generating Data-Driven Mechanistic Hypothesis}
\label{sec:main-results}

Table~\ref{tab:main-results} shows performance across the evaluated agents and AI Scientists on easy and hard tasks. Across both difficulty levels, agent-generated hypotheses achieve lower MCA and higher MCNLL than the QMR and PMR. Codex with GPT-5.6 Sol and scientific skills achieves the highest MCA across the evaluated configurations, but substantial gaps remain relative to the QMR (agent minus QMR, easy/hard: MCA -22.7/-32.1; MCNLL +0.468/+0.785). The gaps are even larger relative to the PMR. Both empty and shuffled controls also perform substantially worse than the MRs on both metrics, supporting the evaluation's validity to distinguish task-relevant mechanistic content from absent or mismatched content. These results indicate substantial room for improvement in AI agents' ability to generate mechanistic hypotheses from the available observations.

\subsection{Evaluating Hypothesis Generation Quality and Efficiency across Harnesses and Skills}
\label{sec:harnesses_and_skills}

\begin{figure}[htbp]
    \centering
        \subfigure[MCA Changes (pp)]{%
            \includegraphics[height=3.5cm]{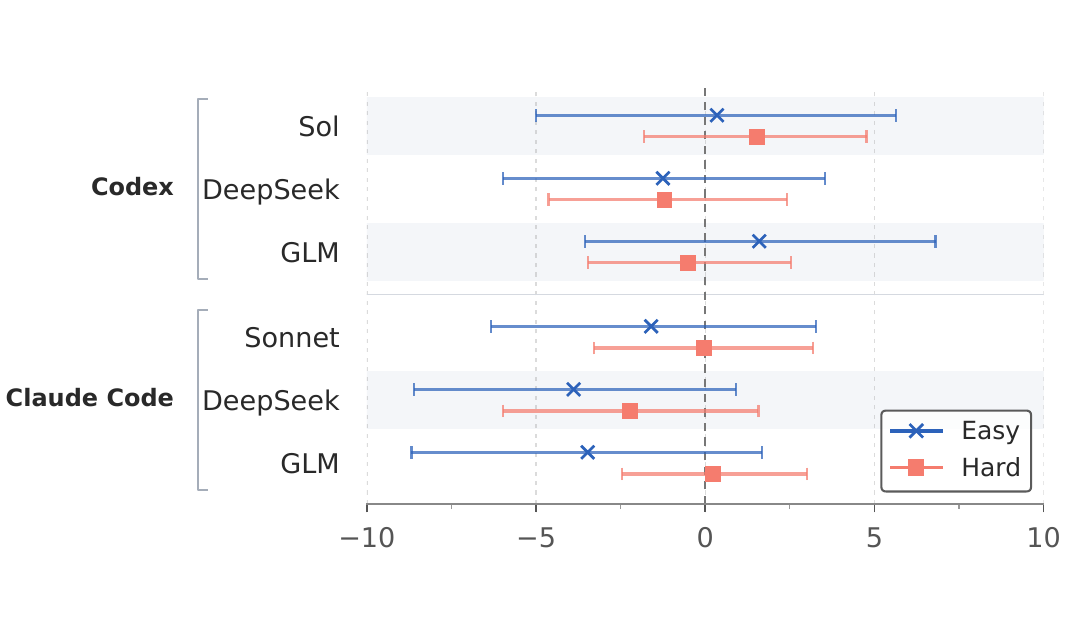}%
        }%
        \hspace{1mm}%
        \subfigure[MCNLL Changes]{%
            \includegraphics[height=3.5cm]{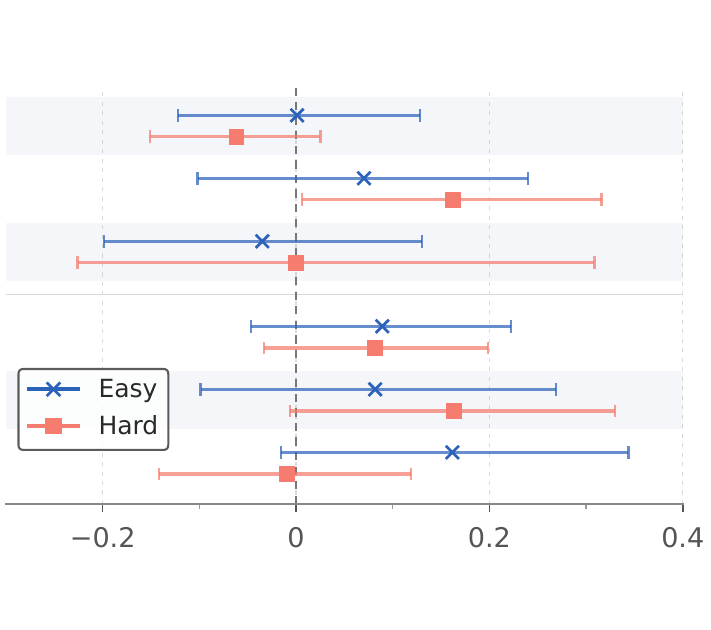}%
        }%
        \hspace{1mm}%
        \subfigure[Runtime Changes (\%)]{%
            \includegraphics[height=3.5cm]{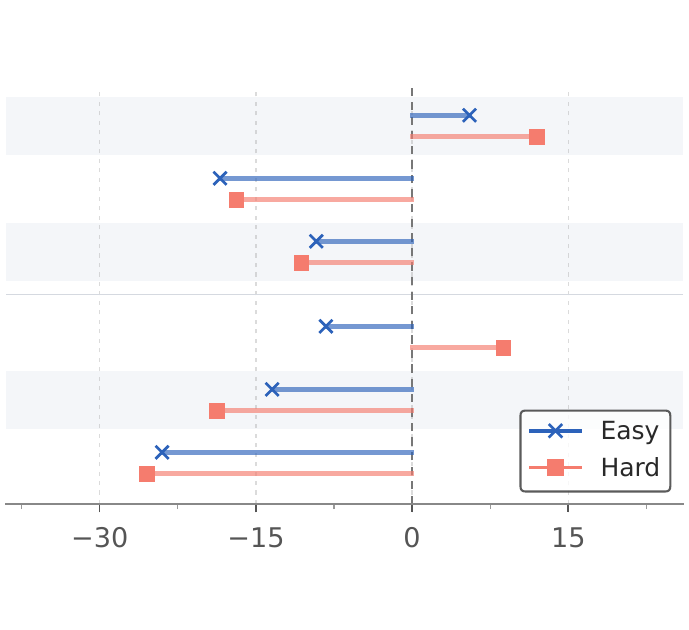}%
        }%
    \vspace{-4mm}
    \caption{Changes associated with scientific-skill access. Points show with-skills minus without-skills differences for easy and hard tasks; horizontal bars denote 95\% confidence intervals. Panels report changes in (a) MCA, (b) MCNLL, and (c) agent execution time.}
    \label{fig:exp_skills}
\end{figure}

Harness choice affects performance even when the model backend and skill setting are held fixed (Figure~\ref{fig:exp_skills}). Claude Code generally achieves higher MCA than Codex with DeepSeek-V4-Flash and GLM-5.2. The clearest advantage appears with DeepSeek-V4-Flash, where both MCA and MCNLL favor Claude Code under both skill settings; without skills, the easy-task MCA difference is 7.8 percentage points (95\% CI: [2.2, 13.3]). These gains accompany a different execution pattern: Claude Code makes more tool calls in all four matched comparisons and takes longer than Codex with DeepSeek-V4-Flash under both skill settings. With GLM-5.2, Claude Code is slightly slower without skills but faster with skills. Thus, the harness influences both hypothesis-generation performance and execution cost.
\vspace{-5mm}

Access to scientific skills reduces execution time for DeepSeek-V4-Flash and GLM-5.2 under both harnesses, but these gains do not consistently translate into better mechanistic hypotheses. Mean execution time decreases by 10.1-18.1\% across these four configurations. However, the pattern is model-dependent: Codex with GPT-5.6 Sol takes longer with skills, whereas Claude Code with Claude Sonnet 5 shows little runtime change. Nevertheless, MCA and MCNLL show limited or inconsistent improvements. Together, these results suggest that making scientific guidance available can improve execution efficiency, but does not consistently improve the mechanistic content of the resulting hypotheses.

\begin{takeaway}[Takeaways]
    \begin{itemize}[leftmargin=1.3em,topsep=1pt,noitemsep]
        \item Harness choice affects both hypothesis quality and execution cost.
        \item Scientific skill access can shorten execution without consistently improving mechanistic hypothesis generation.
    \end{itemize}
    
\end{takeaway}

\subsection{Evaluating Mechanistic Hypothesis Generation in AI Scientists}
\label{sec:aiscientist}

AI-scientist workflows do not consistently improve mechanistic hypothesis generation. AI-Scientist-v2 achieves 59.2\% and 43.8\% MCA on easy and hard tasks, respectively, below Codex with GPT-5.6 Sol. AutoDiscovery performs even lower in the matched 40-task comparison, reaching 27.5\% and 27.1\% MCA, compared with 47.1\% and 35.8\% for Codex with GPT-5.6 Sol without skills. Their MCNLL profiles differ from their MCA rankings, with AutoDiscovery assigning higher probability to the ground-truth direction on the hard subset despite its lower MCA. These results show that incorporating scientific-discovery workflows changes performance profiles but does not by itself improve agents' ability to formulate mechanistic hypotheses from observations.

\begin{takeaway}[Takeaways]
    \begin{itemize}[leftmargin=1.3em,topsep=1pt,noitemsep]
        \item AI-scientist workflows do not consistently improve mechanistic hypothesis generation over a general agent using the same underlying model.
    \end{itemize}
    
\end{takeaway}

\subsection{Evaluating Effects of Evidence Access and Retention}
\label{sec:interventions}

We examine two settings that change how agents access and retain evidence from the public observations. In the \textbf{evidence summaries} setting, agents receive precomputed summaries of how responses vary across conditions in the public data. These summaries make relevant empirical relationships explicit without providing new observations. In the \textbf{structured records} setting, agents receive no precomputed evidence. Instead, they are instructed to record their own findings, the conditions under which each finding holds, its supporting evidence, and potential falsifiers, and to revisit these records before producing their final hypotheses. We evaluate both settings on fixed 40 tasks using Codex with GPT-5.6 Sol and Claude Code with DeepSeek-V4-Flash, the highest-MCA model configurations within their respective harnesses in the main experiments.

\begin{figure}[tbp]%
    \centering 
    \vspace{-2mm}    
     \subfigure[Easy Task / MCA] { \includegraphics[width=0.23\linewidth]{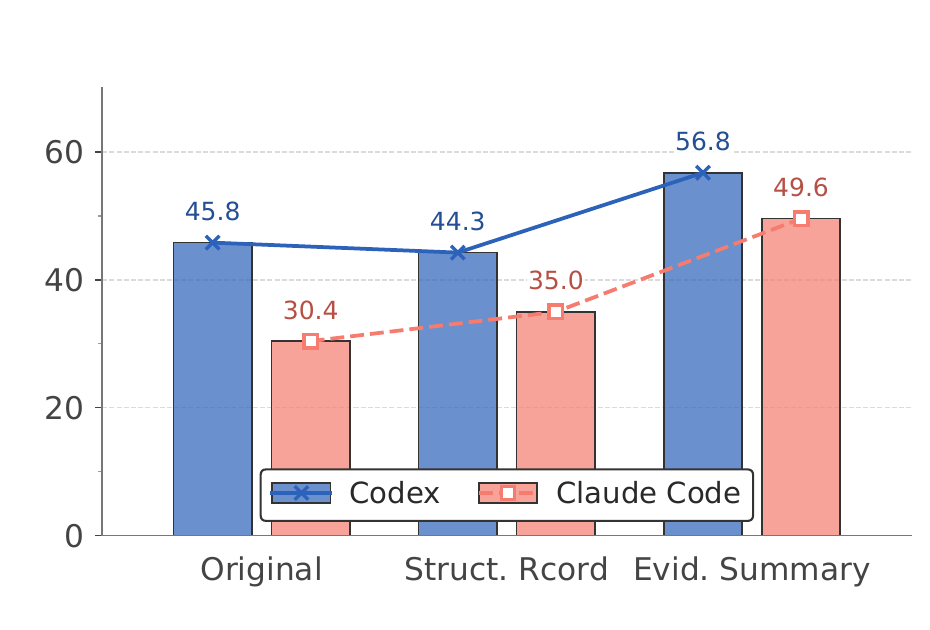} }
     \subfigure[Easy Task / MCNLL]{ \includegraphics[width=0.23\linewidth]{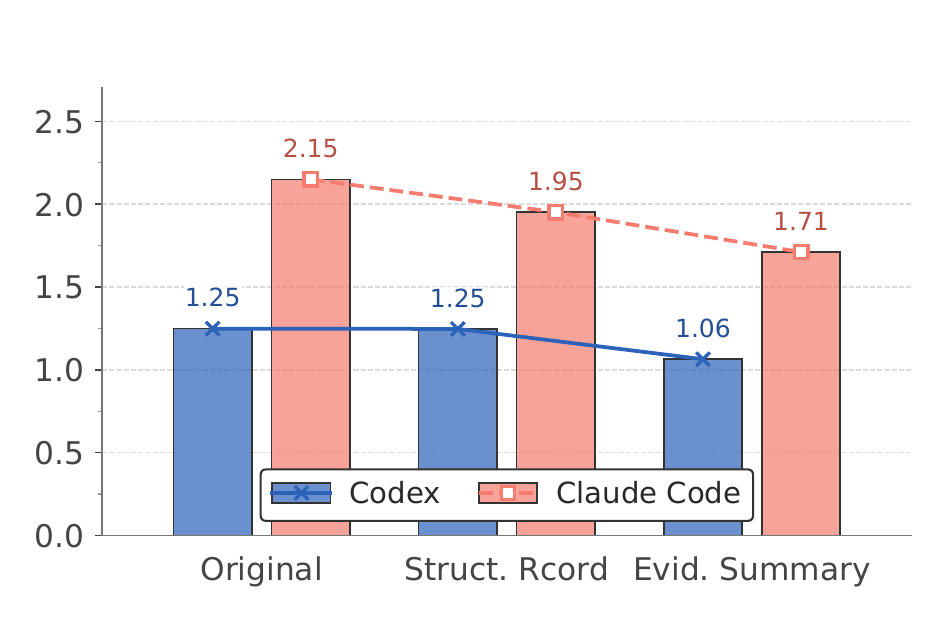} }
     \subfigure[Hard Task / MCA]{ \includegraphics[width=0.23\linewidth]{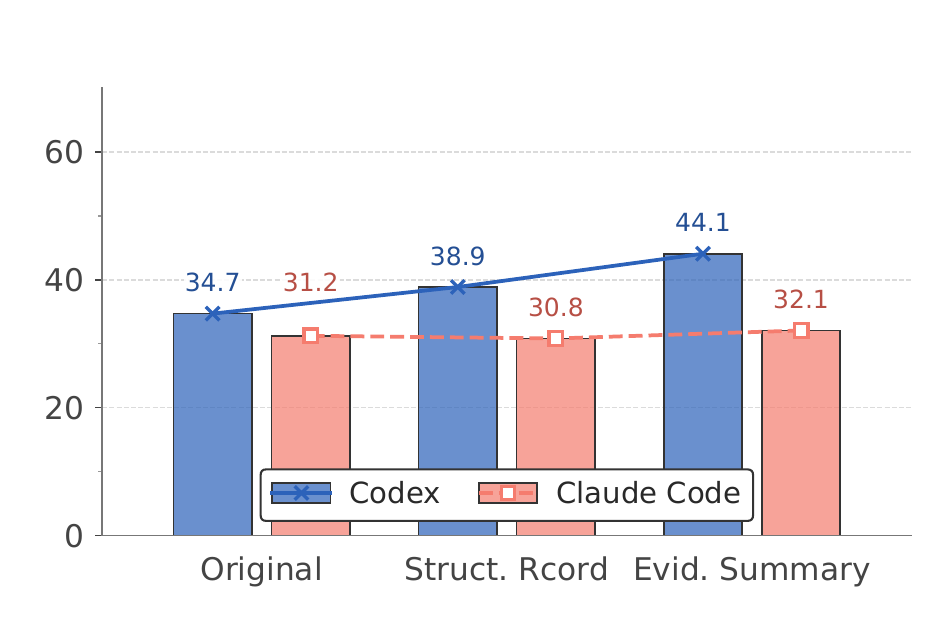} }
      \subfigure[Hard Task / MCNLL]{ \includegraphics[width=0.23\linewidth]{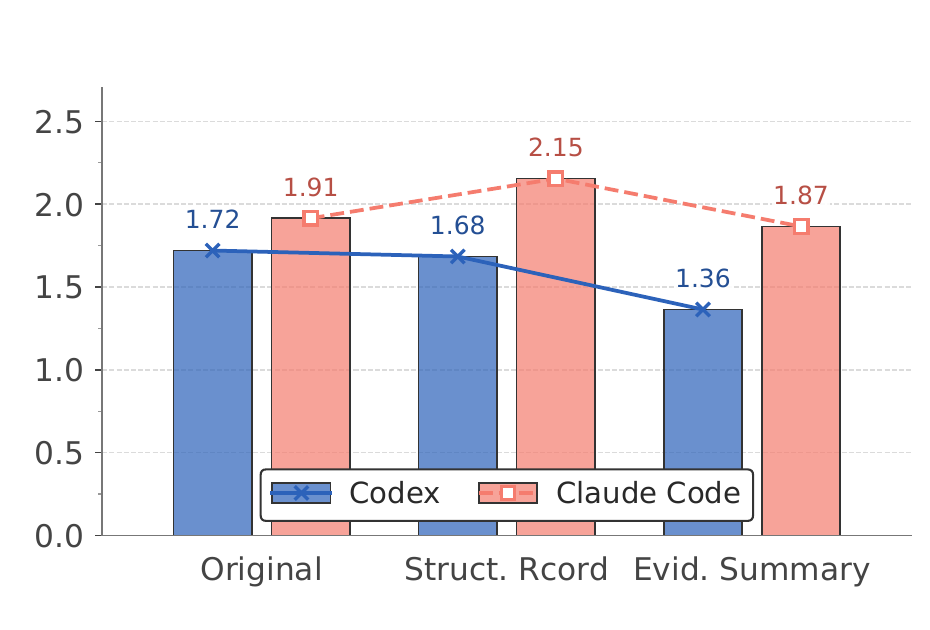} }
    \vspace{-3mm}
    \caption{Interventions on 40 tasks per configuration. Lines connect scores across conditions; missing outputs use the empty control.} 
    \label{fig:exp_interventions}
\end{figure}

Providing evidence summaries improves MCA by 10.0 percentage points for both configurations, with MCNLL reductions of 0.288 for Codex with GPT-5.6 Sol and 0.244 for Claude Code with DeepSeek-V4-Flash (Figure~\ref{fig:exp_interventions}). MCNLL also improves on tasks with accepted outputs in all settings. These results suggest that making empirical relationships explicit can help agents use the available observations to formulate mechanistic hypotheses. Nevertheless, performance remains below the MRs, indicating that supplying these findings does not eliminate the difficulty of inferring the generating process. Structured records yield smaller MCA gains of 1.9 and 2.1 percentage points for Codex with GPT-5.6 Sol and Claude Code with DeepSeek-V4-Flash, respectively, without consistent improvement in MCNLL. For Codex with GPT-5.6 Sol, the resulting hypotheses more accurately imply unchanged responses but less accurately imply increases or decreases.
These results suggest that the difficulty of mechanistic hypothesis generation for AI agents lies partly in discovering and selecting relevant evidence from the available observations. Structured records provide a way to retain findings during extended analysis, but cannot compensate for evidence that agents have overlooked. Their limited gains further suggest that preserving findings and integrating them into a mechanistic hypothesis are distinct demands; support for the former may not fully address the latter.

\begin{takeaway}[Takeaways]
    \begin{itemize}[leftmargin=1.3em,topsep=1pt,noitemsep]
        \item Evidence summaries improve mechanistic hypothesis generation more consistently than structured records, suggesting that agents leave useful evidence in the available observations unexploited.
        \item The limited gains from structured records suggest that retaining findings and using them to formulate a mechanistic hypothesis are distinct demands.
    \end{itemize}
    
\end{takeaway}

%% file: chapters/main_results.tex
\begin{table}[h]
\caption{Performance of mechanistic hypothesis generation evaluation. QMR and PMR denote qualitative and parameterized mechanism references. Empty and Shuffled denote absent and mismatched hypothesis controls.}
\label{tab:main-results}
\centering
\begingroup
\footnotesize
\setlength{\tabcolsep}{2pt}
\renewcommand{\arraystretch}{1.12}
\resizebox{0.85\linewidth}{!}{%
\begin{tabular}{@{}cccccccccc@{}}
\toprule
\multirow{3}{*}{\textbf{Type}}
 & \multirow{3}{*}{\textbf{Models}}
 & \multicolumn{4}{c}{\textbf{Without Skills}}
 & \multicolumn{4}{c}{\textbf{With Skills}} \\
\cmidrule(lr){3-6}\cmidrule(lr){7-10}
 & & \multicolumn{2}{c}{\textbf{MCA} (\%)$\uparrow$}
   & \multicolumn{2}{c}{\textbf{MCNLL} $\downarrow$}
   & \multicolumn{2}{c}{\textbf{MCA} (\%)$\uparrow$}
   & \multicolumn{2}{c}{\textbf{MCNLL} $\downarrow$} \\
\cmidrule(lr){3-4}\cmidrule(lr){5-6}
\cmidrule(lr){7-8}\cmidrule(lr){9-10}
 & & Easy & Hard & Easy & Hard & Easy & Hard & Easy & Hard \\
\midrule
\multirow{4}{*}{\shortstack[c]{Ref. \& Controls}}
 & \emph{PMR}
 & \textbf{96.8} & \textbf{97.5} & \textbf{0.165} & \textbf{0.133}
 & \textbf{96.8} & \textbf{97.5} & \textbf{0.165} & \textbf{0.133} \\
 & \emph{QMR}
 & \textbf{87.6} & \textbf{82.9} & \textbf{0.458} & \textbf{0.538}
 & \textbf{87.6} & \textbf{82.9} & \textbf{0.458} & \textbf{0.538} \\
 & \emph{Empty}
 & 16.7 & 16.8 & 1.935 & 2.002
 & 16.7 & 16.8 & 1.935 & 2.002 \\
 & \emph{Shuffled}
 & 38.1 & 39.9 & 2.916 & 3.146
 & 38.1 & 39.9 & 2.916 & 3.146 \\
\midrule
\multirow{3}{*}{Codex}
 & GPT-5.6 Sol
 & \underline{64.6} & \underline{49.3} & \underline{0.925} & \underline{1.385}
 & \underline{64.9} & \underline{50.8} & \underline{0.926} & \underline{1.323} \\
 & DeepSeek-V4-Flash
 & 41.4 & 31.2 & 1.725 & 2.157
 & 40.1 & 30.0 & 1.795 & 2.319 \\
 & GLM-5.2
 & 40.9 & 28.7 & 1.657 & 2.046
 & 42.5 & 28.1 & 1.622 & 2.046 \\
\midrule
\multirow{3}{*}{Claude Code}
 & Claude Sonnet 5
 & 43.7 & 28.8 & \underline{1.499} & \underline{1.882}
 & 42.1 & 28.8 & \underline{1.588} & \underline{1.964} \\
 & DeepSeek-V4-Flash
 & \underline{49.2} & \underline{34.0} & 1.641 & 2.027
 & \underline{45.3} & \underline{31.8} & 1.723 & 2.190 \\
 & GLM-5.2
 & 45.7 & 28.7 & 1.510 & 2.067
 & 42.2 & 29.0 & 1.672 & 2.057 \\
\midrule
\multirow{2}{*}{AI scientists}
 & AI-Scientist-v2
 & 59.2 & 43.8 & 1.031 & 1.448
 & - & - & - & - \\
 & AutoDiscovery
 & 27.5 & 27.1 & 1.474 & 1.526
 & - & - & - & - \\
\bottomrule
\end{tabular}%
}
\endgroup
\end{table}

%% file: chapters/4_conclusion.tex
\section{Conclusion}
\label{sec:conclusion}

This paper introduces \benchmarkname, a benchmark for data-driven mechanistic hypothesis generation that combines paper-derived mechanisms across 14 fields with independently collected empirical datasets.
We then develop an evaluation framework that assesses open-form mechanistic hypotheses through their consequences under withheld conditions, with ground truth outcomes determined by the same paper-derived mechanism that generates the observations. Through systematic experiments with AI agents, we identify limitations in data-driven mechanistic hypothesis generation and examine how evidence availability and the retention of findings affect the hypotheses they produce.